\documentclass[10pt, conference, compsocconf]{IEEEtran}

\IEEEoverridecommandlockouts
\usepackage{cite}
\usepackage{amsmath,amssymb,amsfonts}
\usepackage{algorithmic}
\usepackage{graphicx}
\usepackage{textcomp}
\usepackage{xcolor}
\usepackage{array}
\usepackage{multirow}
\usepackage{diagbox}
\usepackage{mathrsfs}
\usepackage{ulem}
\usepackage{color}
\usepackage{url}

\def\BibTeX{{\rm B\kern-.05em{\sc i\kern-.025em b}\kern-.08em
    T\kern-.1667em\lower.7ex\hbox{E}\kern-.125emX}}
\begin{document}
\title{EATEN: Entity-aware Attention for Single Shot Visual Text Extraction\\
}

\author{
\IEEEauthorblockN{
			He guo, Xiameng Qin, Jiaming Liu, Junyu Han, Jingtuo Liu, Errui Ding}
		\IEEEauthorblockA{
			Department of Computer Vision Technology(VIS), Baidu Inc.\\
			Email: 
			\{guohe, qinxiameng, liujiaming03, hanjunyu, liujingtuo, dingerrui\}@baidu.com}
			\vspace{-2em}
	}

\maketitle

\begin{abstract}

Extracting entity from images is a crucial part of many OCR applications, such as entity recognition of cards, invoices, and receipts. Most of the existing works employ classical detection and recognition paradigm. This paper proposes an Entity-aware Attention Text Extraction Network called \textbf{EATEN}, which is an end-to-end trainable system to extract the entities without any post-processing. In the proposed framework, each entity is parsed by its corresponding entity-aware decoder, respectively. Moreover, we innovatively introduce a state transition mechanism which further improves the robustness of entity extraction. In consideration of the absence of public benchmarks, we construct a dataset of almost 0.6 million images in three real-world scenarios (train ticket, passport and business card), which is publicly available at \url{https://github.com/beacandler/EATEN}. To the best of our knowledge, EATEN is the first single shot method to extract entities from images. Extensive experiments on these benchmarks demonstrate the state-of-the-art performance of EATEN.
\end{abstract}


\section{Introduction}
\label{introduction}
Recently, scene text detection and recognition, two fundamental tasks in the field of computer vision, have become increasingly popular due to their wide applications such as scene text understanding \cite{milyaev_fast_2015}, image and video retrieval \cite{gomez_single_2018}.
Among these applications, extracting Entity of Interest (EoI) is one of the most challenging and practical problems, which needs to identify texts that belong to certain entities. Taking passport (Fig.  \ref{fig:dataset_sample_images}) for example, there are many entities in the image, such as Country, Name, Birthday and so forth. In practical applications, we only need to output the texts for some predefined entities, \textit{e.g.} ``China'' or ``USA'' for the entity ``Country'', ``Jack'' or ``Rose'' for the entity ``Name''. Previous approaches \cite{TranditionMethod_Invoice_2007, TranditionMethod_Invoice_Icdar2009, Layout_Invoice_ICTSD2018} mainly adopt two steps, in which text information is extracted firstly via OCR (Optical Character Recognition), and then EoIs are extracted by handcrafted rules or layout analysis.

Nevertheless, in detection and recognition paradigm, engineers have to develop post-processing steps, which are handcrafted rules to determine which part of the recognized text belongs to the predefined EoIs. It's usually the post-processing steps, rather than the ability of detection and recognition, restraints the performance of EoIs extraction. For example, if the positions of entities have a slight offset to the standard positions, inaccurate entities will be extracted due to sensitive template representation. 


\begin{figure*}
    \centering
    \includegraphics[scale=0.365]{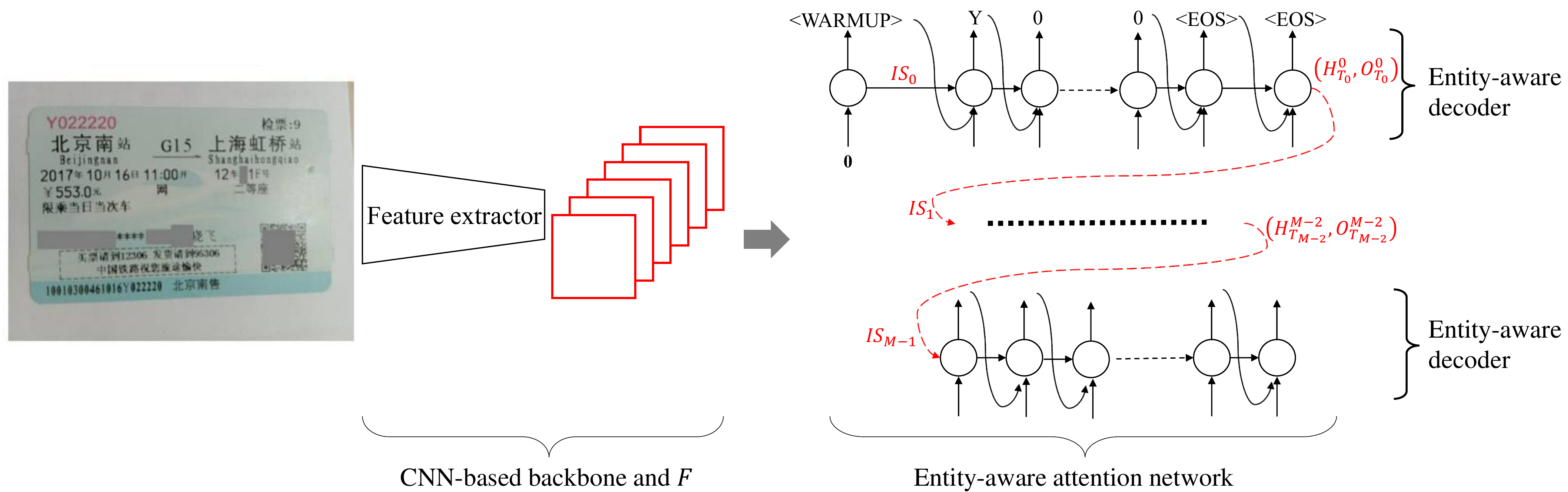}
    \caption{EATEN: an end-to-end trainable framework for extracting EoIs from images. We use a CNN to extract high-level visual representations, denoted as $F$, and an entity-aware attention network to decode all the EoIs. We pad the ground truth of every decoder to a fixed length with $<$EOS$>$ characters. $IS_0$ is the initial state of the 1st entity-aware decoder. The red dash line shows the process of state transition. Taking $D_{M-1}$  and $D_{M-2}$ as example,  the State transition can be expressed as $IS_{M-1}=(H_{0}^{M-1}, O_{0}^{M-1})=(H_{T_{M-2}}^{M-2}, O_{T_{M-2}}^{M-2})$.}
    \label{fig:framework_1}
\end{figure*}

In this paper, a single shot Entity-aware Attention Text Extraction Network (EATEN) is proposed to extract EoIs from images within a single neural network. 
As shown in Fig. \ref{fig:framework_1}, we use a CNN-based feature extractor
to extract feature maps from original image. 
Then we design an entity-aware attention network, which is composed of multiple entity-aware decoders, initial state warm up and state transition between decoders, to capture all entities in the image. 
Compared with traditional methods, EATEN is an end-to-end trainable framework instead of multi-stage procedures. 
EATEN is able to cover most of the corner cases with arbitrary shapes, projective/affine transformations, position drift without any correction due to the introduction of spatial attention mechanism.
Since there are rarely large datasets available for visual EoI extraction, we construct a new dataset with about 0.6 million real and synthetic images, which includes train tickets, passports, and business cards. 
The proportions of Chinese, English and digital EoIs in the dataset are $40.5\%, 21.6\%, 37.8\%$ respectively.
EATEN shows significant improvements compared to other related methods, and it achieves $4.5$, $4.1$, and $2.8$ FPS respectively in three scenarios (\textit{i.e.}, train ticket, passport, and business card) on a Tesla P40 GPU.
The results show that EATEN can deal with EoIs in Chinese as well as those in English and digits.

In summary, our approach has the following three main contributions:  
(1) we propose an end-to-end trainable system to extract EoIs from images in a unified framework. (2) we design an entity-aware attention network with multiple decoders and state transition between contiguous decoders so that the EoIs can be quickly located and extracted without any complicated post-process. (3) we establish a large dataset for visual EoI extraction. Three real-world scenarios are selected to make the dataset much more practical and useful.


\section{Related Work}

Typically, scene text reading \cite{li_show_2018, gupta_synthetic_2016, zhou_east:_2017, liu2018detecting} falls  into two categories, one stream is recognizing all the texts in the image, and the other one is merely recognizing the EoIs, which is also called entity extraction or structural information extraction. 

The first stream of scene text reading usually contains two modules, scene text detection \cite{hu2017wordsup} and scene text recognition \cite{li_show_2018}. A text line is described as a rectangle or a quadrilateral, or even a mask region by using regression or segmentation methods \cite{zhou_east:_2017, xu_textfield:_2018, Zhang2019LookMT}.
After we obtain the location of texts, many recognition algorithms, such as CRNN \cite{shi_end--end_2017} and attention-based methods \cite{li_show_2018}, could be utilized to obtain the texts in the image. Recently, detection and recognition branches are merged into an end-to-end framework, and jointly trained simultaneously \cite{sui_novel_2018}. 

EoIs extraction is the second stream of scene text reading and is vital to real applications like credit card's entities recognition. Classical approaches are based on rules and templates \cite{TranditionMethod_Invoice_Icdar2009, TranditionMethod_DocExtract_ACM2012}, which firstly recognizes all the texts in the image by OCR methods,
and then extracts EoIs with projective transformation, handcrafted strategies and many post-processes. Spatial connections, segmentation and layout analysis methods \cite{TranditionMethod_Invoice_2007, Layout_Invoice_PartBased2016, Layout_Bart2010InformationEB} were employed to extract structural information of EoIs. Gall \textit{et al.} \cite{Rinon_receiptRecg_ACCV18} created an embedding that merged spatial and linguistic features for extracting EoI information from images. However, the processing pipelines are redundant and complex so that rule adjustment should be very carefully. Most of those methods verified their algorithm on self-established datasets which are only for privately experiments. Thus, we construct a real-world scenario dataset.

In addition, many methods \cite{wojna_attention-based_2017, Zisserman_readingtext_IJCV16} were presented to avoid using results of OCR technique. 
Word spotting based methods \cite{gomez_single_2018, Wilkinson_CtrlF_ICCV17} directly predicted both bounding boxes and a compact text representation of the words within them. 
D. He \textit{et al.} \cite{he_guided_2018} verified the existence of a certain text string in an image with spatial attention mechanism.
Although these new methods outperform the traditional methods, the extracted words have no entity information. To solve this problem, we present an entity-aware attention network to extract EoIs for all the entities.

\section{Proposed Method}

Fig. \ref{fig:framework_1} shows the framework of EATEN. 
The CNN-based backbone aims to extract high-level visual features from images, and the entity-aware attention network learns entities layout of images automatically and decodes the content of predefined EoIs by \textbf{entity-aware decoders}. In addition, \textbf{initial state warm-up} and \textbf{state transition} are proposed to improve the performance of our method.

We choose Inception v3 \cite{szegedy_rethinking_2016} as the backbone. Feature maps after backbone are noted as $F$, whose shape is $H \times W \times C$, where $C$ is the number of channels.
Let $E=\{E_0, ..., E_{I-1}\}$ denote the set of entities, where $I$ is the number of entities. $D=\{D_0, ..., D_{M-1}\}$ and $\{T_0, ..., T_{M-1}\}$ respectively denote the entity-aware decoders and the time steps of each entity-aware decoder, where $M$ is the number of entity-aware decoders.
Since one decoder is able to capture one or several EoIs, $M$ is not necessarily equal to $I$. To build the semantic relations between the neighboring EoIs, we employ the last state of previous decoder to initialize the current decoder. We also use initial state warm-up to boost the performance of attention mechanism. 
Considering that every decoder is aware of its corresponding entity/entities, we call this network \textbf{Entity-aware attention network}. EATEN has no explicit text recognition and uses the entity-aware decoder to decode the corresponding EoIs directly. No lexicon is used in this work.

\subsection{Entity-aware Decoder}
\label{Charcter_feature_extractor}

\begin{figure}
    \centering
    \includegraphics[scale=0.29]{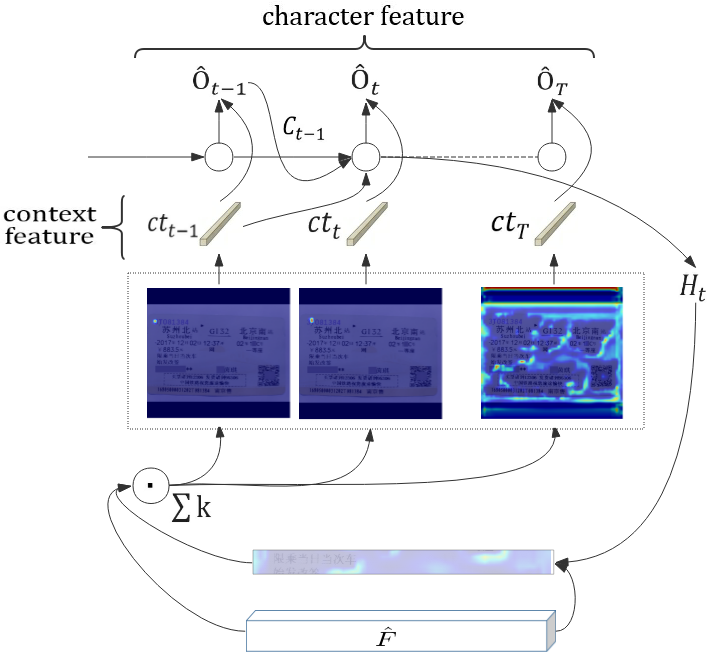}
    \caption{Entity-aware decoder. We use attention mechanism to obtain the context feature, denoted as $ct_t$, and then model the sequences of context feature with an LSTM. At last, we concatenate  $ct_t$ and $O_t$ as character feature, denoted as $\hat{O_t}$. }
    \label{fig:framework_2}
    \vspace{-1.5em}
\end{figure}

In general, the decoders are arranged from left to right and top to bottom. We will assign multiple entities to a decoder if these entities always show up in the same text line, such as SS/TAN/DS in Fig. \ref{fig:cases}. The decoding process for a single entity is illustrated in Fig. \ref{fig:framework_2}.
In each decoding step, the entity-aware decoder firstly uses entity-aware attention mechanism to obtain the corresponding context feature. The context feature, combined with previously predicted character, is further fed into an LSTM unit as input. And then the LSTM will update context feature and predict current character. Given a hidden state $H_t$ from the LSTM and feature maps $F$, context feature $ct_t$ is computed as
\vspace{-0.5em}
\begin{equation}
\label{equation_1}
e ^ t _ { k } =  V  \circ \tanh \left( W _ { h } \times H _ { t } + W _ { f } \times \hat{F _ { k}} \right)
\end{equation}
\vspace{-1.5em}
\begin{equation}
\label{equation_3}
ct_t = \sum _ {k = 1} ^ {K}  \left( \frac 
{e_k^t} {\sum_{i=1}^{K} e_i^t} \circ \hat{F} _ {k} \right),\ K = H \times W
\vspace{-0.6em}
\end{equation}
where $\hat{F}$ is rearranged from $F$, and its size is $(H \times W,\ C)$. $W_f$ is accomplished by a $1\times1$ convolution. $\circ$ is elementwise multiplication and $\times$ is linear transformation. $W_h$, $W_f$ and $V$ are weight matrices to be learned. 

The LSTM updates $H_{t-1}$ to $H_t$, and its output is combined with an updated context feature $ct_t$ to generate final predicted distribution over characters $P(y _ t | y _ {1: t-1}, F)$. 
\begin{equation}
\vspace{-1.5em}
\label{equation_6}
( H_t, O_t ) = LSTM(W _ { c } \times C _ { t - 1 } + W _ {ct _ {1}}  \times ct _ { t - 1 }, H _ { t - 1 })
\end{equation}

\begin{equation}
\vspace{-1.5em}
\label{equation_7}
 \hat{O}  _ { t } =  W _ { o } \times O _ { t } + W _ { ct _ { 2 } } \times ct _ { t } 
\end{equation}

\begin{equation}
\label{equation_output}
    P(y_ t | y _ {1: t-1}, F) = softmax(\hat{O}_t) 
\end{equation}
where $C _ { t - 1 }$ is the one-hot encoding of previous character. $W _ c$， $W _ {ct _ {1}}$, $W _ o$ and $W _ {ct _ {2}}$ are the weight matrices to be learned. $y_ t$ is the predicted character of $t$ time step. The decoder index $m$ is omitted here for simplicity.

\subsection{Initial State Warm-up}

In our observation, the vanilla entity-aware decoding model can not converge. The main problem lies in the first decoding step of the first entity-aware decoder. The first decoding step possibly generates identical character no matter what images are fed to EATEN. The decoder requires information from $H_{t-1}$, $C_{t-1}$, and $ct_{t-1}$ to update context feature $ct_{t}$ and predict a character in step $t$. However, in the first step $t=1$, $H_{0}$, $C_{0}$, and $ct_{0}$ are initialized by constant values, only the updated $ct_{1}$ is utilized to predict $y_1$. It's barely possible for the network to make a successful prediction from $ct_{1}$ only. To make the first step prediction stable, a buffer decoding step is introduced to warmup the decoder. The buffer decoding step outputs a spacial character token $<$WARMUP$>$ that will be discarded later, and the hidden state is computed as
\vspace{-0.6em}
\begin{equation}
\label{equation_8}
( H^0_1,\ O^0_1 ) = LSTM \left(W _ {ct _ {1}}  \times \sum _ {k = 1} ^ {K}  ( \frac {e_k^0} {\sum_{i=1}^{K}
e_i^0
} \circ \hat{F} _ {k} )\right)
\end{equation}

As shown in Fig. \ref{fig:framework_1}, this hidden state is regarded as the initial state of the 1st entity-aware decoder, denoted as $IS_0 = (H^0_1,\ O^0_1)$. 

\begin{figure}
    \centering
    \includegraphics[scale=0.3]{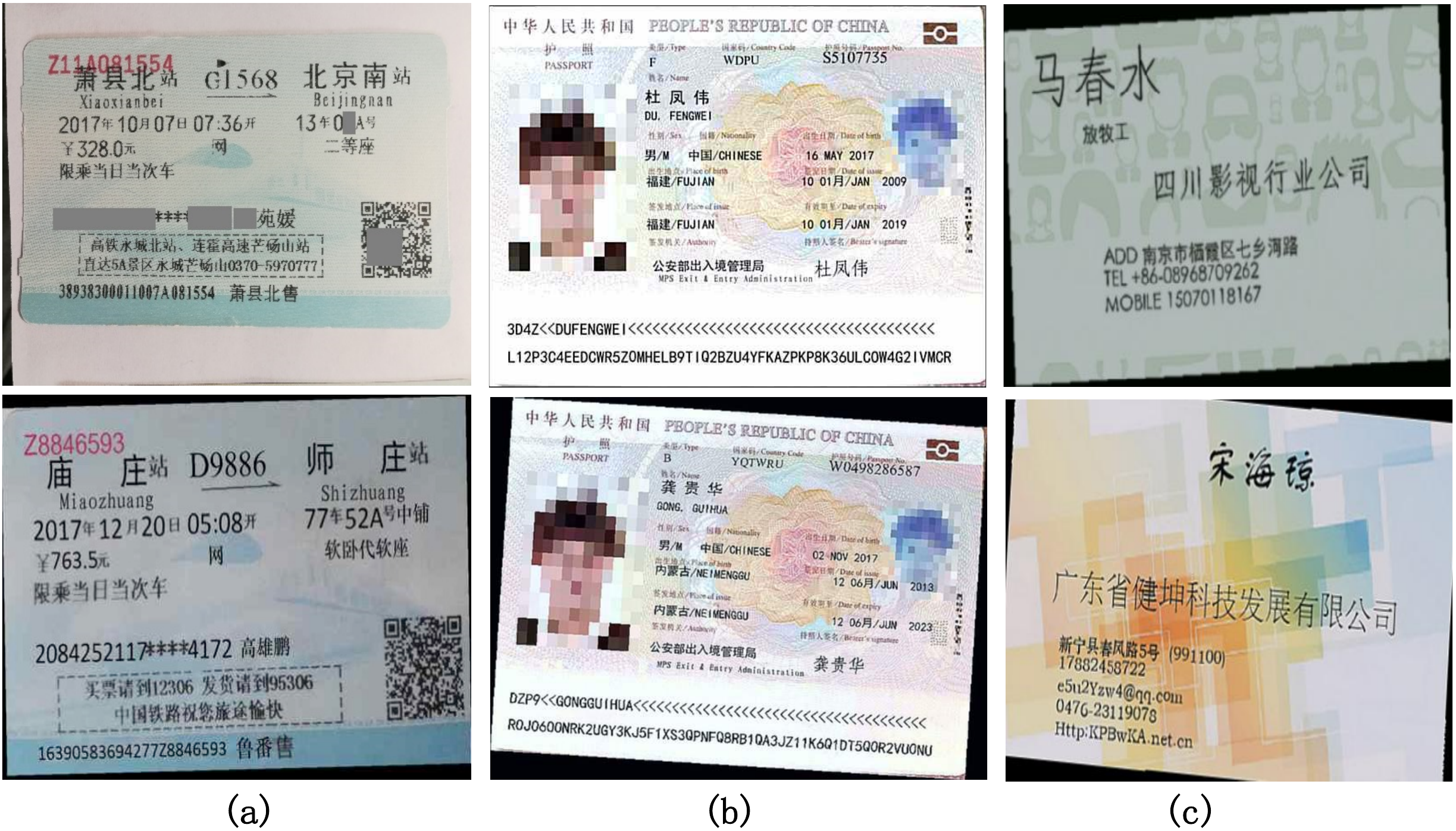}
    \caption{Sample images: (a) The first row shows the real image of train ticket and the second row demonstrates the synthetic image of train ticket. (b) synthetic images of passport. (c) synthetic images of business card.}
    \label{fig:dataset_sample_images}
    \vspace{-1.2em}
\end{figure}

\subsection{State Transition}

The decoders are independent of each other in our entity-aware attention network. In most of scenarios, the EoIs usually have strong semantic relations. Thus, we design state transition between neighboring decoders to establish the relations of the EoIs.
As can be seen in Fig. \ref{fig:framework_1}, assuming that ($H_{T_m}^m$, $O_{T_m}^m$) is the last state of  $D_m$. The initial state of its next decoder $D_{m+1}$ is expressed as $IS_{m+1}=(H_{0}^{m+1}, O_{0}^{m+1})=(H_{T_m}^m, O_{T_m}^m)$.
Therefore, the semantic information of all the previous decoders will be integrated into the current decoder. Experimental results show that state transition improves the performance obviously.

The model is trained by maximum likelihood estimation, and the overall loss is expressed as
\vspace{-0.6em}
\begin{equation}
\begin{aligned}
    l = & \sum_{m=0}^{M-1} l^m = \sum_{m=0}^{M-1} \left( -log P(y^m|\hat F) \right) \\
      = & \sum_{m=0}^{M-1} \left( -\sum_{t=1}^{T_m} log P (y^m_t | y_{1:t-1}^m, \hat F) \right)
\vspace{-0.6em}
\end{aligned}
\end{equation}
Where $l^m$ is the loss for the $m$-th decoder and is defined as the negative log likelihood.

\section{Dataset}

\subsection{Synthetic Data Engine}

The text information in 
many applications always
 contains various 
personal 
information, such as ID Card Number, Name, Home Addresses, \textit{etc.}, which must be erased before releasing to the public. 
Data synthesis is a way to bypass the privacy problem, and it also shows great help in scene text detection and scene text recognition \cite{gupta_synthetic_2016, jaderberg_synthetic_2014}.
Following the success of synthetic word engine \cite{jaderberg_synthetic_2014}, we propose a 
more efficient 
text rendering pipeline.

\begin{figure}
    \centering
    \includegraphics[scale=0.32]{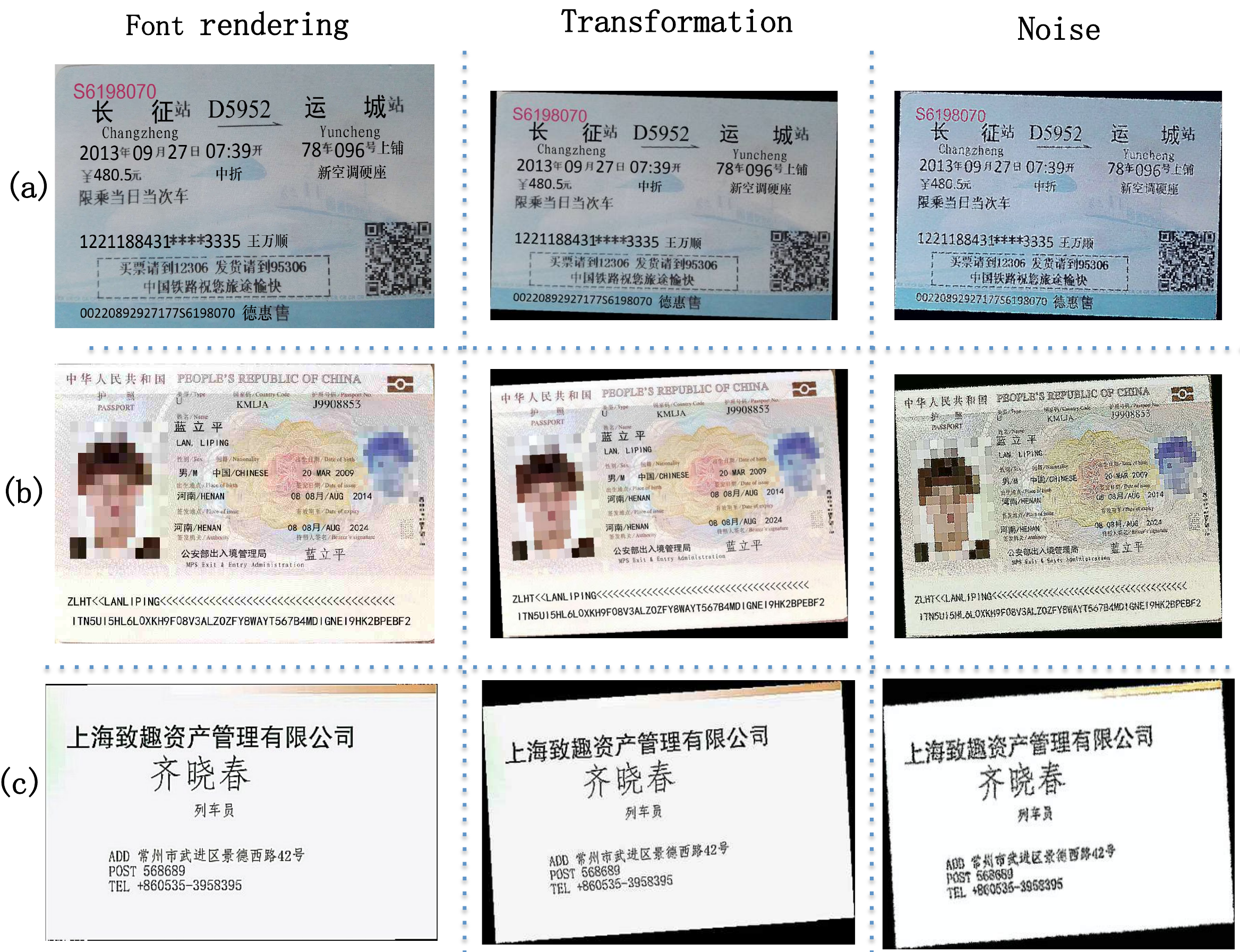}
    \caption{Synthetic image samples: (a) train ticket. (b) passport. (c) business card. 
    From left to right: images after font rendering, transformation, and noising.}
    \label{fig:dataset_synth_process}
\end{figure}

Fig. \ref{fig:dataset_synth_process} illustrates the synthetic image generation processes and displays some synthetic images. The generation processes contain four steps:
\begin{itemize}
    \item \textit{Text preparing.}
    To make the synthetic images more general, we collected a large corpus including Chinese name, address, \textit{etc.} by crawling from the Internet.
    \item \textit{Font rendering.}  We select one font for a specific scenario, and the EoIs are rendered on the background template images using an open image library. Especially, in the business card scenario, we prepared more than one hundred template images containing 85 simple background and pure images with random color to render text. 
    \item \textit{Transformation.}  We rotate the image randomly in a range of [-5, +5] degree, then resize the image according to the longer side. Elastic transformation is also employed.
    \item \textit{Noise.}  Gaussian noise, blur, average blur, sharpen, brightness, hue, and saturation are applied. 
\end{itemize}

The images are resized and padded to $512\times512$ for train ticket and $600\times450$ for passport. The template images of business card have already been fixed to $500\times300$ so that we do not need any extra resizing and padding operations. The aspect ratio of text is preserved in these three scenarios, which is important for text recognition. 
The images and their corresponding labels of EoIs are saved, simultaneously. There's no overlap between EoIs of the test set and EoIs of the training set. 

\subsection{ Benchmark Datasets}

\textbf{Train ticket.}
Sample images are shown in Fig. \ref{fig:dataset_sample_images}.
As we can see, the top left corner is the real image of train ticket. The real images were shot in a finance department with inconsistent lighting conditions, orientations, background noise, imaging distortions, \textit{etc}. Then we remove the private information, such as First Name, ID card number, QR code, and Seat Number. 
Meanwhile, synthetic images were also produced with the synthetic data engine. 
The train ticket dataset includes a total of 2k real images and 300k synthetic images, 0.4k real images of which are used for testing and 
the rest for training (90\%) and validation (10\%).

\textbf{Passport.}
As shown in Fig. \ref{fig:dataset_sample_images}(b), we select one template image for passport, and erase the private information, such as Name, Passport Number, Face, on this image. 
The passport dataset includes a total 100k synthetic images, 2k images of which are used for testing and the rest for training and validation.

\textbf{Business card.}
The synthetic images for business card are shown in Fig. \ref{fig:dataset_sample_images}(c), the positions of the EoIs are not constant and some EoIs may not exist, which is a challenge for extracting EoIs.
The business card dataset includes a total 200k synthetic images, 2k images of which are used for testing and the rest for training and validation.

In contrast of train ticket, we only verify our approach in the synthetic dataset for passport and business card due to privacy concern.

\section{Experimental Results}
\subsection{Implementation Details}
We use SGD (stochastic gradient descent) optimizer with initial learning rate of  0.04, exponential learning rate decay factor of 0.94 for every 4 epoch and momentum of 0.9. 
To prevent overfitting, we regularize the model using weight decay 1e-5 in Inception v3 and character softmax layer, label smoothing with epsilon 0.1 and the values of LSTM Cell state clipping to [-10, 10]. In order to make the training process more stable, we clip the gradient by norm with 2.0. We use 8 Tesla P40 GPUs to train and 1 Tesla P40 GPU to test our model. Batch size for each dataset is different due to the different input image size, the number of entity-aware decoders and decoding steps of each decoder.

\begin{table*}[htbp]
\label{result_train_ticket}
\caption{\upshape Results on train ticket. We verify our method on three datasets including only synthetic data, only real data, and fused data. The evaluation metric is mEA.}
\centering
\begin{tabular}{ |c|c|c|c|  }
 \hline
 \multirow{2}{*}{\diagbox{Method}{Dataset}}&
 Only synth &
 Only real &
 Fused  \\ \cline{2-4}&
 270k synth &
 1.5k real &
 270k synth  + 1.5k real\\
 \hline
 General OCR   &
 79.2\% &
 79.2\% &
 79.2\%\\
 \hline
 Attention OCR\cite{wojna_attention-based_2017} &
 26.0\% &
 68.1\% &
 90.0\%\\
 \hline
 EATEN w/o state   &
 57.0\% &
 55.4\% &
 91.0\% \\
 \hline
 EATEN   &
 55.1\% &
 86.2\% &
 95.8\% \\
 \hline
\end{tabular}
\vspace{-2em}
\end{table*}

\subsection{Experiment Setting}

\textbf{Evaluation Metrics.} In train ticket and passport scenario, we define mean entity accuracy (mEA) to benchmark EATEN, which can be computed as 
\vspace{-0.8em}
\begin{equation}
    \label{MEA}
    mEA = \sum_{i=0}^{I-1}\mathbb I(y^i==g^i)/ I
\end{equation}
where $y^i$ is the predicted text and $g^i$ is the target text of the $i$-th entity. $I$ is number of of entities and $\mathbb I$ is the indicator function that return 1 if $y^i$ is equal to $g^i$ or 0 otherwise.
In business card scenario, not all EoIs are guaranteed to appear, we define mean entity precision (mEP), mean entity recall (mER), and mean entity F-measure (mEF) to benchmark our task, which can be computed as
\vspace{-1.2em}
\begin{equation}
    \label{MEP}
    mEP = \sum_{i=0}^{I_{p}-1}\mathbb I(y^i==g^i)/I_{p}
\end{equation}
\vspace{-0.6em}
\begin{equation}
    \label{MER}
    mER = \sum_{i=0}^{I_{g}-1}\mathbb I(y^i==g^i)/I_{g}
\end{equation}
where $I_{p}$ is the number of non-null predicted entities, $I_{g}$ is the number of non-null target entities. 
When prediction and target are all null, indicator function returns 0.
mEF is the harmonic average of mEP and mER.

\begin{table}[t] 
\label{result_passport_business}
\caption{\upshape Results on passport and business car.}
\centering
\begin{tabular}{ |c|c|c|c|c|  }
 \hline
 \multirow{3}{*}{\diagbox{Method}{Scenario}}&
 Passport &
 \multicolumn{3}{|c|}{Business card} \\ \cline{2-5}&
 96k synth &
 \multicolumn{3}{|c|}{196k synth} \\  \cline{2-5}
 & mEA & mEP & mER & mEF \\
 \hline
 General OCR
 & 11.1\%$^{\mathrm{1}}$
 & 59.9\%
 & 60.5\%
 & 60.2\%\\
 \hline
 Attention OCR\cite{wojna_attention-based_2017}
& 81.6\%
& 79.5\%
& 79.2\%
& 79.4\%\\
 \hline
 EATEN w/o state
& 84.9\%
& 89.7\%
& 89.7\%
& 89.7\%\\
 \hline
 EATEN
& 90.8\%
& 90.0\%
& 89.6\%
& 89.8\%\\
\hline
\multicolumn{5}{l}{$^{\mathrm{1}}$
Due to the entity words (Name, Passport No., \textit{etc.}) are extremely }\\
\multicolumn{5}{l}{blurred, the matching rules were totally missed.}\\
\multicolumn{5}{l}{}
\end{tabular}
\vspace{-2em}
\end{table}

\textbf{Entity-aware attention networks.} We design different entity-aware attention networks for three scenarios.
Generally speaking, we regard a group of EoIs that have strong semantic or position relations as a block, and use an entity-aware decoder to capture all the EoIs in this block. 
In train ticket scenario, we introduce five decoders to capture eight EoIs, the decoding steps, which refers to how many characters at most one decoder could generate of each decoder are set to 14, 20, 14, 12, 6 respectively. The decoding steps of each decoder are decided by the max number of characters of its corresponding EoIs. The EoIs that assigned to the decoders are Ticket Number (TCN), (Starting Station (SS), Train Number (TAN),  Destination Station (DS)), Date (DT), (Ticket Rates (TR), Seat Category (SC)), and Name (NM).
In passport scenario, we design five decoders to cover seven EoIs, decoding steps of each decoder are 25, 5, 15, 35, 35. The EoIs assigned to the decoders are Passport Number, Name, (Gender, Birth Date), Birth Place, (Issue Place, Expiry Date). Nine decoders are set to cover ten EoIs for business card, and decoding steps of each decoder are 21, 13, 21, 21, 21, 21, 32, 10, 21. The EoIs of each decoder are Telephone, Postcode, Mobile, URL, Email, FAX, Address, (Name, Title), Company. If an entity-aware decoder is responsible of capturing more than one EoIs, it  
generates an $<$EOS$>$ token in the end of decoding each EoI. Different EoIs are separated by the $<$EOS$>$ tokens.

\textbf{Compared methods} We compare several baseline methods with our approach: (1)
General OCR. A typical paradigm, OCR and matching, that firstly detects and reads all the text by OCR engine\footnote{http://ai.baidu.com/tech/ocr/general}, and then extracts EoIs if the content of text fits predefined regular expressions or the position of text fits in designed templates.
(2) Attention OCR\cite{wojna_attention-based_2017}. It reads multiple lines of scene text by attention mechanism and has achieved state-of-the-art performance in several datasets. We adapt it to transcribe the EoIs sequentially, using $<$EOS$>$ tokens to separate different EoIs. (3) EATEN without state transition. This method is for ablation study, to verify the efficiency of proposed state transition.

\subsection{Performance}

We report our experimental results in this section.
In train ticket scenario, as we can see from the 4th column of Table I, EATEN shows significant improvement over General OCR, Attention OCR, and EATEN w/o state (16.6\%, 5.8\%, and 4.8\%).

The results of passport can be seen in the 2nd column of Table II, EATEN obtains a huge improvement over all other methods (especially General OCR, 90.8\% v.s. 11.1\%). 
Assuming that “entity words” are the texts that always appeared in the images, fixedly. We find that the blurred parts for passport are “entity words”, which are the exact words “Name” or “Passport No.” etc. Those entity words are used to locate corresponding EoIs content value in template matching. If entity words are missed, template matching will fail. However, since EATEN does not rely on entity words, it could generalize on these scenarios as long as the EoIs are recognizable.
we can also see from the 5th column of Table II, EATEN significantly outperforms the General OCR by relative 29.6\% in business card scenario. 

Meanwhile, benefiting from the simplified overall pipeline, EATEN achieves high speed performance.
As we can see in Table III, EATEN respectively achieves $\textbf{7}\times$, $\textbf{12}\times$, $\textbf{4}\times$ speedups over General OCR. Compared with Attention OCR, which also use attention mechanism, EATEN shows its advantage (the last two rows).
\begin{table}[t] 
\label{time-cost}
\caption{\upshape Time costs of EATEN on three scenarios.}
\centering
\begin{tabular}{ |c|c|c|c|c|  }
 \hline
 \diagbox{Method}{Scenario}&
 Train ticket & Passport & Business card \\
 \hline
 General OCR
 & 1532ms
 & 3000ms
 & 1600ms\\
 \hline
 Attention OCR\cite{wojna_attention-based_2017}
& 335ms
& 260ms
& 428ms\\
 \hline
 EATEN
& 221ms
& 242ms
& 357ms\\
 \hline
\end{tabular}
\vspace{-1.5em}
\end{table}

All experiments of the three scenarios also illustrate the generalization of EATEN. 
The layout of entities in train ticket and passport scenario is fixed, yet is inconstant for business card. The EoIs of business card can appear in arbitrary positions.
Combining the performances of EATEN on three scenarios, we conclude that EATEN can not only cover real-world applications which have a fixed layout, but also applications whose layout is flexible. It can be used 
in many real-world applications including card recognition (\textit{e.g.}, ID card, driving license) and finance invoice recognition (quota invoice, air ticket).
Considering that these results are released without any parameters tuning and post-process, there is room for improvement.

\subsection{Discussion}
\textbf{Synthetic data engine.}
We conduct some experiments to explore the impact of the synthetic data engine.
As shown in the 2nd column of Table I, EATEN trained with images generated by our synthetic data engine can achieve 55.1\% in mEA, which illustrates that synthetic images are capable of simulating real image to some degree. 
Considering the data distribution of the synthetic data has a large gap with the real data, 
it is reasonable that EATEN w/o state outperforms EATEN slightly (57.0\% v.s. 55.1\%), which can be considered as a regular fluctuation. 
We can also see from the 3rd column of Table I, the performance of EATEN trained on only real data is relatively poor compared with trained on fused data. Actually, the number of real data is too small so that the model may be overfitted. 
We add extra synthetic data to the real data and train a new model. As shown in the 4th column of Table I, 
the results show that model trained with fused train data outperforms model trained with only real data by a large margin (95.8\% v.s. 86.2\%, 91.0\% v.s. 55.4\%, 90.0\% v.s. 68.1\%).
In summary, synthetic images is capable of simulating real-world images to some degree and makes contribution to the improvement of final performance.
\begin{figure}
    \centering
    \includegraphics[scale=0.4]{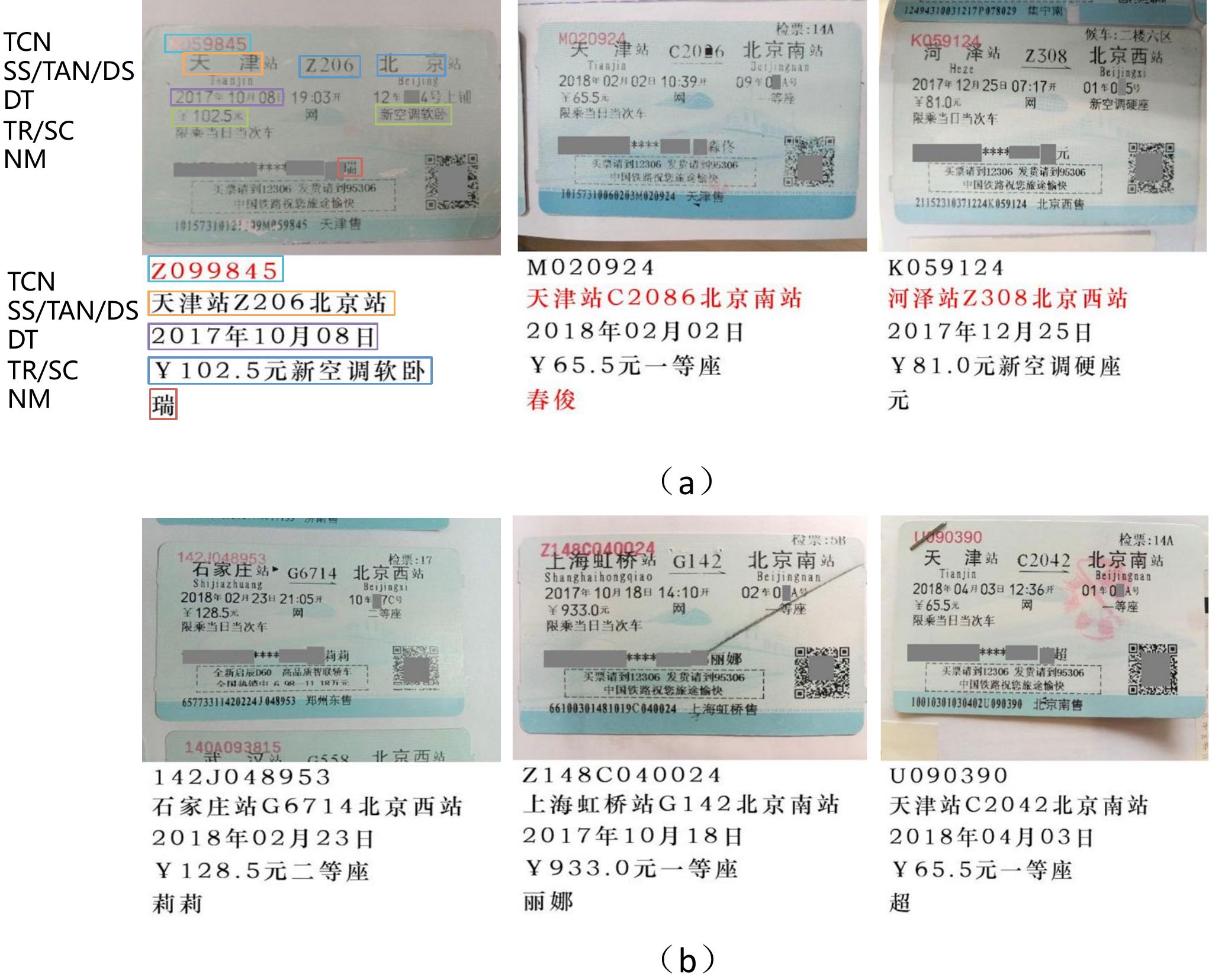}
    \caption{Examples of EoIs extraction: red font indicates EoI is wrongly predicted and black font indicates EoI is correctly predicted. (a) bad cases. (b) good cases. Entry names are labeled for better demonstration.}
    \label{fig:cases}
\end{figure}

\textbf{Examples.}
We analyze 0.4k real images to understand the weakness and advantage of EATEN. As shown in Fig. \ref{fig:cases}, The most common error is composed of three parts including challenging lighting conditions, unexpected disturbance, and serious text fusion, which are also challenging in traditional OCR. 
This problem can be alleviated by adding a denoising component to EATEN. On the other hand, we can also see some good cases from Fig. \ref{fig:cases}, EATEN can cover most of the texts with arbitrary shapes, projective/affine transformations, position drift without any correction. In some situations, such as text fusion and slight lighting, EATEN can also extract EoIs correctly, which shows its robustness in complex situations.

\section{conclusions}
In this paper, we proposed an end-to-end framework called EATEN for extracting EoIs in images. A dataset with three real-world scenarios were established to verify the efficiency of the proposed method and to complement the research of EoI extraction. 
In contrast to traditional approaches based on text detection and text recognition, 
EATEN is efficiently trained without bounding box and full text annotations, and directly predicts target entities of an input image in one shot without any bells and whistles. 
It shows superior performance in all the scenarios, and shows full capacity of extracting EoIs from images with or without a fixed layout. 
This study provides a new perspective on text recognition, EoIs extraction, and structural information extraction. 

{
\bibliographystyle{IEEEtran}
\normalem
\bibliography{im2fields}
}

\end{document}